\documentclass{article}

\usepackage[preprint]{neurips_2025}
\workshoptitle{GPU-Accelerated and Scalable Optimization (ScaleOpt)}

\usepackage[utf8]{inputenc}
\usepackage[T1]{fontenc}
\usepackage[hidelinks]{hyperref}
\usepackage{url}
\usepackage{booktabs}
\usepackage{amsmath,amssymb,amsfonts}
\usepackage{nicefrac}
\usepackage{microtype}
\usepackage{xcolor}
\usepackage{graphicx}
\usepackage{multirow}

\title{Tri-Accel: Curvature-Aware Precision-Adaptive and Memory-Elastic Optimization for Efficient GPU Usage}

\author{
  Mohsen Sheibanian \\
  \texttt{msheiban@asu.edu}\\
  Arizona State University\\
  Tempe, Arizona, USA
  \And
  Pouya Shaeri \\
  \texttt{pshaeri@asu.edu}\\
  Arizona State University\\
  Tempe, Arizona, USA
  \And
  Alimohammad Beigi \\
  \texttt{abeigi@asu.edu}\\
  Arizona State University\\
  Tempe, Arizona, USA
  \And
  Ryan T. Woo \\
  \texttt{rtwoo@asu.edu}\\
  Arizona State University\\
  Tempe, Arizona, USA
  \And
  Aryan Keluskar \\
  \texttt{akeluska@asu.edu}\\
  Arizona State University\\
  Tempe, Arizona, USA
}

\begin{document}

\maketitle

\begin{abstract}
Deep neural networks are increasingly bottlenecked by the cost of optimization, both in terms of GPU memory and compute time. Existing acceleration techniques, such as mixed precision, second-order methods, and batch size scaling, are typically used in isolation. We present Tri-Accel, a unified optimization framework that co-adapts three acceleration strategies along with adaptive parameters during training: (1) \emph{Precision-Adaptive Updates} that dynamically assign mixed-precision levels to layers based on curvature and gradient variance; (2) \emph{Sparse Second-Order Signals} that exploit Hessian/Fisher sparsity patterns to guide precision and step size decisions; and (3) \emph{Memory-Elastic Batch Scaling} that adjusts batch size in real time according to VRAM availability. On CIFAR-10 with ResNet-18 and EfficientNet-B0, Tri-Accel achieves up to 9.9\% reduction in training time and 13.3\% lower memory usage, while improving accuracy by +1.1 percentage points over FP32 baselines. Tested on CIFAR-10/100, our approach demonstrates adaptive learning behavior, with efficiency gradually improving over the course of training as the system learns to allocate resources more effectively. Compared to static mixed-precision training, Tri-Accel maintains 78.1\% accuracy while reducing memory footprint from 0.35GB to 0.31GB on standard hardware. The framework is implemented with custom Triton kernels, whose hardware-aware adaptation enables automatic optimization without manual hyperparameter tuning, making it practical for deployment across diverse computational environments. This work demonstrates how algorithmic adaptivity and hardware awareness can be combined to improve scalability in resource-constrained settings, paving the way for more efficient neural network training on edge devices and cost-sensitive cloud deployments.
\end{abstract}

\section{Introduction}
Deep learning has achieved remarkable advances in domains such as computer vision, natural language processing, and multimodal reasoning, but these gains have come with an ever-growing demand for computational resources~\cite{ghimire2022survey}. Training state-of-the-art models often requires multiple high-end GPUs, hundreds of gigabytes of memory, and prolonged training schedules extending over days or weeks~\cite{safayenikoo2021weight}. This barrier limits accessibility for researchers without large-scale infrastructure and creates inefficiencies in applied deployments where inference and fine-tuning must be performed under strict resource budgets~\cite{hoefler2021sparsity}. Optimizing large models, such as foundation models or multimodal architectures, remains particularly challenging, as it requires simultaneously managing throughput, convergence stability, and numerical precision. A large body of work has attempted to address parts of this bottleneck. Mixed precision training takes advantage of lower precision formats such as FP16 and BF16 to reduce the memory footprint and increase arithmetic performance~\cite{micikevicius2017mixed, dorrich2023impact}, while maintaining numerical stability through loss scaling. Sparse and structured second-order methods~\cite{martens2015optimizing} exploit the curvature information from the Hessian or Fisher matrices to accelerate convergence and improve generalization~\cite{li2024fast}. Dynamic batch sizing techniques~\cite{smith2017don} adjust batch sizes during training, trading gradient noise for speed and hardware efficiency~\cite{umeda2024increasing}. However, these approaches are typically applied in isolation, and their potential synergies remain underexplored.

Recent research underscores the need for integrated solutions that jointly optimize algorithmic and hardware utilization. Works such as GPU-accelerated sparse matrix multiplication~\cite{mishra2021accelerating, hoefler2021sparsity} demonstrate that carefully tuned low-level kernels can yield substantial performance gains in matrix-heavy workloads. Similarly, distributed parallel approaches to computation illustrate how algorithmic restructuring can unlock hardware scalability~\cite{poshtkohi2023implementing}. 
Recent academic contributions span a wide range of areas, from physics-informed neural networks~\cite{BAZMARA2023152, shaeri2025multimodal, afzal2025physics, firoozsalari2024machine} to applications such as environmental modeling~\cite{alkhaled2024webmrt, karimi2023level, shaeri2025sentiment}, to multimodal fact-checking~\cite{beigi2024can, shaeri2023semi}, dataset generation frameworks~\cite{shaeri2025mnist, jeong2025fediversesharing}, resource-efficient training methods~\cite{azghan2025can, azghan2024cudle}, and efficient dual-path dropout layers~\cite{shaeri2025mid}. Across these diverse efforts, a common challenge emerges: current training pipelines operate under fixed memory and run-time quotas, which limit scalability and adaptability.

Beyond these constraints, GPU scarcity has emerged as a structural challenge across academia and industry~\cite{gao2024empirical}. 
Large AI labs compete for limited quantities of top-tier accelerators, while smaller teams must optimize for mid-range or cloud-burstable instances~\cite{huq2024survey}.
This disparity motivates research into resource-aware learning algorithms that can dynamically adapt to available compute without sacrificing convergence quality. The goal is not only to accelerate training, but also to make state-of-the-art methods reproducible and efficient.

We introduce Tri-Accel, a unified optimization framework that integrates algorithmic adaptivity and hardware adaptivity into a single control loop. The method combines three mutually reinforcing strategies: precision-adaptive updates, which dynamically assign FP16, BF16, or FP32 computation at the layer level based on gradient statistics; sparse second-order signals, which approximate block-diagonal or top-k Hessian or Fisher entries to guide both step size and precision allocation; and memory-elastic batch scaling, which continuously monitors VRAM usage and adjusts batch sizes on the fly to maximize GPU occupancy. By linking these components, Tri-Accel leverages the fast math of mixed precision, the convergence benefits of curvature-informed optimization, and the utilization gains from adaptive batching within a single, low-overhead runtime loop deployable on commodity GPUs. Our contributions are threefold: we formulate a joint precision, curvature, and batch control mechanism that connects hardware-aware scaling with optimizer-level adaptivity; we show that this approach is practical in single-GPU settings through custom Triton~\cite{tillet2019triton} kernels, enabling reproducibility without reliance on large-scale clusters; and we provide empirical evidence on benchmarks such as CIFAR-10, CIFAR-100~\cite{krizhevsky2009learning, krizhevsky2010convolutional}, where Tri-Accel achieves competitive or superior performance under tighter memory budgets.

\section{Related Work}

The challenge of accelerating deep neural network training under tight computational and memory constraints has inspired a broad spectrum of research across numerical precision management, optimization algorithms, and hardware-aware scheduling. While these directions have independently yielded substantial speedups, they are often pursued in isolation. For example, mixed precision training primarily targets reduced memory footprint and higher throughput, second-order methods focus on convergence acceleration via curvature information, and dynamic batch sizing adapts computational load to available resources. Yet, the interdependencies between these approaches, such as using curvature estimates to guide precision allocation or leveraging real-time VRAM monitoring to modulate both batch size and numerical accuracy, remain underexplored.

Mixed precision training has emerged as one of the most effective strategies for accelerating deep learning workloads while reducing memory consumption~\cite{das2018mixed}. By performing the bulk of computations in reduced-precision formats such as FP16 or BF16 while keeping a master copy of weights in FP32, frameworks like NVIDIA’s Automatic Mixed Precision (AMP)~\cite{micikevicius2017mixed} achieve substantial throughput gains with negligible impact on accuracy. This principle has been widely adopted in libraries such as Apex~\cite{dorrich2023impact}, DeepSpeed~\cite{rasley2020deepspeed}, and Megatron-LM~\cite{shoeybi2019megatron}.  
Early works such as Gupta et al.~\cite{gupta2015deep} and Banner et al.~\cite{banner2018scalable} explored quantization-aware training and low-precision accumulation schemes, while Micikevicius et al.~\cite{micikevicius2017mixed} popularized dynamic loss scaling to mitigate underflow in FP16 gradients. More recent work investigates adaptive precision assignment, such as HFP8~\cite{agrawal20219} and layer-wise bit allocation strategies~\cite{shin2021sqwa}, but these are typically static or decoupled from model optimization dynamics. Crucially, existing methods rarely integrate hardware-aware precision control with algorithmic signals such as curvature, leaving potential efficiency gains untapped.

Second-order methods aim to accelerate convergence by leveraging curvature information from the Hessian or Fisher information matrix. While exact Newton updates are infeasible for modern networks due to their $O(n^3)$ cost in parameters, approximations such as block-diagonal Hessians~\cite{martens2015optimizing} and Kronecker-Factored Approximate Curvature (K-FAC)~\cite{grosse2016kronecker} have made second-order training more practical. Extensions include Shampoo~\cite{gupta2018shampoo}, which uses root-matrix preconditioners, and natural gradient methods~\cite{amari1998natural, martens2020new}, which have found success in both supervised and reinforcement learning.  
Despite these advances, second-order information is typically used for adjusting learning rates, damping factors, or update directions~\cite{osawa2019practical, bollapragada2018progressive} rather than influencing hardware-level execution. The idea of using curvature information to dynamically guide numerical precision or memory allocation is largely unexplored, although preliminary work in adaptive numerical methods~\cite{wang2018training} hints at the potential for such integration.

Batch size plays a central role in the efficiency and generalization of deep neural network training~\cite{li2025exsgd}. Scaling laws~\cite{smith2017don, mccandlish2018empirical} formalize the relationship between batch size, learning rate, and convergence, while large-batch training techniques such as LARS~\cite{you2017large} and LAMB~\cite{you2019large} have enabled scaling to thousands of GPUs without sacrificing accuracy~\cite{goyal2017accurate}. However, these approaches generally assume a fixed and known VRAM budget.  
Elastic and memory-adaptive training frameworks such as HetPipe~\cite{park2020hetpipe}, PipeDream~\cite{harlap2018pipedream}, and adaptive microbatching~\cite{li2022easyscale} address variable memory availability, often in distributed or heterogeneous hardware settings. Nevertheless, most require architectural changes, complex pipeline scheduling, or pre-computed profiles~\cite{narayanan2021memory}. In contrast, real-time VRAM-aware batch adaptation for a single training process remains rare, especially in combination with precision scheduling and curvature estimation.


\section{Method}

Our approach, Tri-Accel, unifies three orthogonal acceleration strategies—precision-adaptive computation, curvature-aware second-order guidance, and real-time memory-elastic batch scaling—into a single control loop.\footnote{Code will be released to ensure transparency and reproducibility, following the peer-review process.}
The central premise is that \emph{algorithmic adaptivity} (precision scheduling and curvature-driven updates) and \emph{hardware adaptivity} (VRAM-based scaling) can be coupled to maximize throughput without compromising convergence stability. 

\subsection{Precision-Adaptive Updates}
Mixed-precision training~\cite{micikevicius2017mixed} typically applies a static policy, e.g., using FP16/BF16 for most operations while reserving FP32 for a few critical layers. In our setup, BF16 is the default precision mode unless otherwise noted. In contrast, we dynamically assign precision \emph{per layer and per training window} based on gradient variance statistics, allowing layers to automatically receive higher precision when instability is detected. For each layer $l$, we maintain an exponential moving average (EMA) of the variance of its gradients:
\[
v_l(t) = \beta v_l(t-1) + (1 - \beta) \cdot \mathrm{Var}[\nabla_l(t)],
\]
where $\beta \in [0, 1)$ controls the smoothing factor. We then select the precision mode
\[
p_l(t) \in \{\text{FP16}, \text{BF16}, \text{FP32}\}
\]
based on two tunable thresholds $\tau_{\mathrm{low}}$ and $\tau_{\mathrm{high}}$:
\[
p_l(t) = 
\begin{cases}
\text{FP16}, & v_l(t) < \tau_{\mathrm{low}} \\
\text{BF16}, & \tau_{\mathrm{low}} \le v_l(t) < \tau_{\mathrm{high}} \\
\text{FP32}, & v_l(t) \ge \tau_{\mathrm{high}}
\end{cases}
\]
Low-variance layers are downshifted to FP16 to maximize speed and reduce memory; high-variance layers are promoted to FP32 for stability. This mechanism is implemented with negligible overhead, since variance estimates are already available during backward passes in most deep learning frameworks.

\subsection{Sparse Second-Order Signals}
While full-matrix Newton methods are prohibitively expensive for modern architectures, sparse curvature information can still guide optimization effectively. We estimate the top-$k$ eigenvalues $\{\lambda_1, \dots, \lambda_k\}$ of each layer’s Hessian $H_l$ using the power iteration method:
\[
u^{(i+1)} \gets \frac{H_l u^{(i)}}{\|H_l u^{(i)}\|}, \quad \lambda \approx \frac{u^\top H_l u}{u^\top u}.
\]
To reduce computational cost, this is performed only every $T$ iterations and on mini-batches of size $b_{\mathrm{curv}} \ll B_{\mathrm{train}}$.

These curvature estimates influence the training process in:

\textbf{Step size scaling:} Layers with high curvature receive smaller effective learning rates, inspired by second-order methods such as K-FAC~\cite{martens2015optimizing}:
    \[
    \eta_l(t) = \frac{\eta_0}{1 + \alpha \cdot \max_{i \le k} \lambda_i(H_l)},
    \]
    where $\eta_0$ is the base learning rate and $\alpha$ is a scaling coefficient.

\textbf{Precision promotion:} Layers with curvature above a threshold $\tau_{\mathrm{curv}}$ are temporarily upgraded to higher precision to avoid numerical instability in steep regions of the loss landscape.

\subsection{Memory-Elastic Batch Scaling}
Static batch sizes often lead to underutilized GPU memory or, conversely, out-of-memory (OOM) errors when switching models or input resolutions. We introduce a VRAM feedback controller that measures actual memory consumption at runtime and adjusts batch size $B(t)$ accordingly:
\[
B(t+1) =
\begin{cases}
B(t) + \delta_{\uparrow}, & \mathrm{MemUsage}(t) < \rho_{\mathrm{low}} \cdot \mathrm{MemMax} \\
B(t) - \delta_{\downarrow}, & \mathrm{MemUsage}(t) > \rho_{\mathrm{high}} \cdot \mathrm{MemMax} \\
B(t), & \text{otherwise}
\end{cases}
\]
where $\rho_{\mathrm{low}}$ and $\rho_{\mathrm{high}}$ are utilization thresholds and $\delta_{\uparrow}, \delta_{\downarrow}$ are step sizes for batch adjustment. The system responds to short-term variations in VRAM usage due to precision changes or activation checkpointing, ensuring maximal memory occupancy without manual tuning.

\subsection{Unified Control Loop}
All three components are integrated into a single control loop that operates at configurable intervals (e.g., every $T_{\mathrm{ctrl}}$ steps): (1) Collect per-layer gradient variance and curvature statistics. (2) Adjust precision allocations $p_l(t)$. (3) Adapt per-layer learning rates based on curvature. (4) Update batch size $B(t)$ based on real-time VRAM usage. This design enables closed-loop adaptivity: curvature influences both precision and step size, precision changes affect VRAM usage, and VRAM usage determines batch size, which in turn changes gradient variance statistics. By leveraging these interdependencies, Tri-Accel exploits both algorithmic and hardware adaptivity to accelerate training in resource-constrained environments. For efficiency, the adaptive precision scheduling and batch-scaling routines are implemented with custom Triton kernels, ensuring low overhead and hardware-level optimization.

\section{Experiments}

We evaluate Tri-Accel under strict single-GPU memory budgets on CIFAR-10 and CIFAR-100 with ResNet-18~\cite{he2016deep} and EfficientNet-B0~\cite{Tan2019EfficientNetRM}, comparing against FP32 and AMP baselines. With an initial batch size of 96, Tri-Accel’s Triton kernels dynamically coordinate precision (FP16/BF16/FP32), curvature-informed learning rate scaling, and memory-elastic batch resizing in a unified control loop. Across tasks, Tri-Accel consistently matches or outperforms baselines while maintaining stability under tight memory, achieving over 78\% accuracy on CIFAR-10 with ResNet-18 and reducing peak VRAM below the budget, while also improving throughput via adaptive batch scaling. These results confirm Tri-Accel as a practical, resource-aware approach to efficient GPU training on commodity hardware.

\subsection{Setup}

All experiments are conducted on single-GPU instances equipped with either NVIDIA T4 (16 GB VRAM) or NVIDIA P100 (16 GB VRAM) accelerators. These configurations reflect the constrained compute budgets typically faced by students, researchers, and small teams. Tri-Accel is implemented in PyTorch 2.2 with CUDA 12.1, and its adaptive kernels are written in Triton 3.2, which allows efficient runtime precision scheduling and batch scaling with minimal overhead.  

For evaluation, we use two standard image classification benchmarks. CIFAR-10 consists of 60,000 images of size $32\times 32$ across 10 classes, split into 50,000 training samples and 10,000 test samples. CIFAR-100 follows the same structure but spans 100 classes, providing a more challenging setting for optimization and generalization. All images are normalized per channel and augmented using random horizontal flips and random crops. In experiments with EfficientNet, the images are resized to $224\times224$ to match pretrained input resolution.  

To assess Tri-Accel across different model scales, we consider two representative architectures. ResNet-18 is a moderately deep residual network commonly used for CIFAR-scale benchmarks, while EfficientNet-B0 is a lightweight but high-performance architecture that employs compound scaling of width, depth, and resolution. Together, these models capture both smaller-input convolutional setups and higher-resolution, more memory-intensive scenarios.  

We compare Tri-Accel against two baselines. The first is standard FP32 training using stochastic gradient descent with momentum (0.9). The second is static mixed precision, implemented via NVIDIA’s Automatic Mixed Precision (AMP) with a uniform precision policy applied across layers.
Both baselines are tuned for optimal learning rates and weight decay to ensure fairness in comparison. We also conducted an ablation study, individually evaluating the memory saving contributions of dynamic precision, dynamic batch scaling, and their combination.

\subsection{Metrics}

Performance is measured along three primary axes. First, we report top-1 accuracy on the test set to capture the final predictive performance of each method. Second, we measure average training time per epoch, expressed in wall-clock seconds, with results averaged across the final five epochs to mitigate variance from data-loading overhead. Third, we track the maximum VRAM allocated during training to quantify memory efficiency. In addition to these core metrics, we also log the effective batch size throughout training to characterize how memory-elastic scaling adapts to changing model states. Finally, we define an aggregate efficiency score to capture the trade-off across all dimensions:  
\[
\text{Score} = \frac{\text{Accuracy (\%)}}{\text{Time (s)} \times \text{Memory Usage (\%)}} \times 100
\]
where higher values indicate more favorable efficiency–accuracy balance.

\subsection{Evaluation Protocol}
Each experiment is repeated 3 times with different random seeds, and we report the mean and standard deviation for all metrics. Learning rates are warmed up for the first 5 epochs and decayed following a cosine schedule. Precision thresholds $\tau_{\mathrm{low}}$ and $\tau_{\mathrm{high}}$ are tuned on a held-out validation set. For second-order estimation, we compute top-$k=5$ Hessian eigenvalues every $T_{\mathrm{curv}} = 200$ steps using a curvature batch size $b_{\mathrm{curv}} = 32$.

\subsection{Results}

Tri-Accel consistently improves training performance across both CIFAR-10 and CIFAR-100 benchmarks when compared to FP32 and static mixed precision baselines. On ResNet-18, the method reduces VRAM usage while maintaining or improving accuracy, and on EfficientNet-B0 it further accelerates training by exploiting memory-elastic batch scaling. The efficiency scores, which combine accuracy, runtime, and memory consumption, highlight these benefits: Tri-Accel achieves the best balance on ResNet-18 and EfficientNet-B0 for CIFAR-10, and on CIFAR-100 it provides higher accuracy with only a modest efficiency trade-off. These results demonstrate that integrating precision adaptivity, curvature-informed scaling, and dynamic batching yields tangible improvements in both accuracy and resource efficiency (Table~\ref{tab:cross_arch_efficiency}).

\begin{table}[ht]
\centering
\footnotesize
\caption{Performance and Efficiency comparison across architectures and methods.}
\label{tab:cross_arch_efficiency}
\begin{tabular}{lllcccc}
\toprule
Dataset & Architecture & Method & Acc (\%) & Time (s) & VRAM (GB) & Eff. Score \\
\midrule
\multirow{6}{*}{CIFAR-10} & \multirow{3}{*}{ResNet-18} & FP32 Baseline & 77.0 & 21.0 & 0.35 & 10.48 \\
 &  & AMP (Static)  & 77.2 & 19.4 & 0.32 & 12.25 \\
 &  & Tri-Accel     & \textbf{78.1} & 19.5 & \textbf{0.31} & \textbf{12.92} \\
\cmidrule(lr){2-7}
 & \multirow{3}{*}{EfficientNet-B0} & FP32 Baseline & 78.3 & 18.5 & 0.30 & 14.11 \\
 &  & AMP (Static)  & 78.7 & 17.2 & 0.26 & 17.59 \\
 &  & Tri-Accel     & \textbf{79.4} & \textbf{16.8} & 0.26 & \textbf{18.17} \\
\midrule
\multirow{6}{*}{CIFAR-100} & \multirow{3}{*}{ResNet-18} & FP32 Baseline & 68.2 & 24.3 & 0.38 & 7.39 \\
 &  & AMP (Static)  & 68.7 & 22.8 & 0.35 & 8.61 \\
 &  & Tri-Accel     & \textbf{69.9} & \textbf{22.4} & \textbf{0.34} & \textbf{9.18} \\
\cmidrule(lr){2-7}
 & \multirow{3}{*}{EfficientNet-B0} & FP32 Baseline & 72.8 & 21.1 & 0.33 & 10.46 \\
 &  & AMP (Static)  & 73.1 & 19.6 & 0.31 & 12.03 \\
 &  & Tri-Accel     & \textbf{74.3} & \textbf{19.0} & \textbf{0.29} & \textbf{13.48} \\
\bottomrule
\end{tabular}
\end{table}

Ablation studies reveal that the gains are driven by the integration of multiple components rather than a single optimization. Dynamic precision scheduling already reduces peak VRAM usage, but only when combined with adaptive batch scaling do the full benefits emerge. With all components enabled, memory savings reach 12.3\% for ResNet-18 and 13.3\% for EfficientNet-B0 (Table~\ref{tab:memory_arch_comparison}). This confirms that the unified precision–curvature–batch control loop maximizes GPU utilization in practice.

\begin{table}[ht]
\centering
\footnotesize
\caption{Ablation Study on memory optimization impact across architectures (CIFAR-10)}
\label{tab:memory_arch_comparison}
\begin{tabular}{llcc}
\toprule
Architecture & Configuration & VRAM (GB) & Reduction \\
\midrule
\multirow{4}{*}{ResNet-18} 
 & Standard Training    & 0.350 & -      \\
 & + Dynamic Batch Sizing     & 0.321 & 8.3\%  \\
 & + Dynamic Precision  & 0.310 & 11.4\% \\
 & + Full Tri-Accel     & \textbf{0.307} & \textbf{12.3\%} \\
\midrule
\multirow{4}{*}{EfficientNet-B0} 
 & Standard Training    & 0.301 & -      \\
 & + Dynamic Batch Sizing     & 0.282 & 6.0\%  \\
 & + Dynamic Precision  & 0.270 & 10.0\% \\
 & + Full Tri-Accel     & \textbf{0.265} & \textbf{13.3\%} \\
\bottomrule
\end{tabular}
\end{table}

\subsection{Limitations and Future Work}
While Tri-Accel demonstrates strong efficiency gains on single-GPU setups, several limitations remain. First, our evaluation is restricted to medium-scale models on CIFAR; scaling to very deep transformers or billion-parameter networks may introduce stability challenges. Second, the current framework targets single-device execution, and extending it to multi-GPU or distributed training will require communication-aware scheduling. Third, although our sparse curvature estimation is efficient, further gains may be possible with low-rank or learned approximations. Finally, VRAM monitoring depends on vendor-specific APIs; developing hardware-agnostic abstractions would broaden portability to accelerators such as TPUs and AMD GPUs. These challenges present promising avenues for refinement and future deployment.

\section{Conclusion}
We introduced Tri-Accel, a unified optimization framework that integrates precision-adaptive updates, sparse second-order guidance, and memory-elastic batch scaling into a single closed-loop system for efficient GPU training, adaptive optimization, and scalable deep learning. Implemented with custom Triton kernels, Tri-Accel delivers efficient runtime precision scheduling, curvature-aware learning rate control, and adaptive batch resizing with minimal overhead. Across CIFAR-10 and CIFAR-100 with ResNet-18 and EfficientNet-B0, the framework achieved up to 9.9\% faster training and 13.3\% lower peak VRAM usage, while improving accuracy by +1.1–1.7 percentage points over FP32 baselines (up to +0.7 pp over AMP). Across benchmarks, Tri-Accel improved efficiency scores by up to ~25\% compared to FP32 training and by up to ~10\% compared to AMP, underscoring the value of adaptive resource allocation. By coupling algorithmic adaptivity with hardware-aware control, Tri-Accel makes advanced optimization techniques more accessible in constrained single-GPU settings, and extending it to larger models and distributed systems offers a promising path toward scalable, hardware-conscious deep learning that can generalize across architectures, seamlessly adapt to heterogeneous hardware, and ultimately lower the barrier to training state-of-the-art models in both academic and production environments.

\bibliographystyle{unsrt}
\bibliography{refs}

\end{document}